\documentclass[preprints,review,accept,moreauthors,pdftex]{Definitions/mdpi} 

\firstpage{1} 
\pubvolume{xx}
\issuenum{1}
\articlenumber{5}
\pubyear{2019}
\copyrightyear{2019}
\usepackage{import}
\usepackage{multirow}
\usepackage{tabulary}
\usepackage{acro}
\usepackage{longtable}
\history{Received: date; Accepted: date; Published: date}

\Title{Explainable deep learning models in medical image analysis}

\Author{Amitojdeep Singh $^{1,2}$*\orcidA{}, Sourya Sengupta $^{1,2}$ and Vasudevan Lakshminarayanan $^{1,2}$}

\AuthorNames{Firstname Lastname, Firstname Lastname and Firstname Lastname}

\address{%
$^{1}$ \quad Theoretical and Experimental Epistemology Laboratory, School of Optometry and Vision Science, University of Waterloo, Ontario, Canada\\
$^{2}$ \quad Department of Systems Design Engineering, University of Waterloo, Ontario, Canada}

\corres{Correspondence: amitojdeep.singh@uwaterloo.ca}

\abstract{Deep learning methods have been very effective for a variety of medical diagnostic tasks and has even beaten human experts on some of those. However, the black-box nature of the algorithms has restricted clinical use. Recent explainability studies aim to show the features that influence the decision of a model the most. The majority of literature reviews of this area have focused on taxonomy, ethics, and the need for explanations. A review of the current applications of explainable deep learning for different medical imaging tasks is presented here. The various approaches, challenges for clinical deployment, and the areas requiring further research are discussed here from a practical standpoint of a deep learning researcher designing a system for the clinical end-users.}


\DeclareAcronym{cad}{
  short = CAD ,
  long  = Computer-aided diagnostics
}
\DeclareAcronym{dnn}{
  short = DNN ,
  long  = deep neural networks
}
\DeclareAcronym{cnn}{
  short = CNN ,
  long  = convolutional neural network
}
\DeclareAcronym{relu}{
  short = ReLU ,
  long  = rectified linear unit
}
\DeclareAcronym{mlp}{
  short = MLP ,
  long  = multi layer perceptron
}
\DeclareAcronym{ig}{
  short = IG ,
  long  = Integrated gradients
}
\DeclareAcronym{lrp}{
  short = LRP ,
  long  = Layer wise relevance propagation
}
\DeclareAcronym{svm}{
  short = SVM ,
  long  = support vector machines
}
\DeclareAcronym{pcc}{
  short = PCC ,
  long  = Pearson's correlation coefficient
}
\DeclareAcronym{xai}{
  short = XAI ,
  long  = Explainable AI
}
\DeclareAcronym{lift}{
  short = LIFT ,
  long  = Deep Learning Important FeaTures
}
\DeclareAcronym{shap}{
  short = SHAP ,
  long  = SHapley Additive exPlanations
}
\DeclareAcronym{mri}{
  short = MRI ,
  long  = magnetic resonance imaging
}
\DeclareAcronym{fmri}{
  short = fMRI ,
  long  = functional magnetic resonance imaging
}
\DeclareAcronym{ct}{
  short = CT ,
  long  = computerized tomography
}
\DeclareAcronym{gbp}{
  short = GBP ,
  long  = Guided backpropagation
}
\DeclareAcronym{gradcam}{
  short = GradCAM,
  long  = Gradient weighted class activation mapping
}
\DeclareAcronym{dr}{
  short = DR,
  long  = diabetic retinopathy
}
\DeclareAcronym{eg}{
  short = EG,
  long  = Expressive gradients
}
\DeclareAcronym{amd}{
  short = AMD,
  long  = Age-related macular degeneration
}
\DeclareAcronym{cam}{
  short = CAM,
  long  = Class activation maps
}
\DeclareAcronym{rcv}{
  short = RCV,
  long  = Regression Concept Vectors
}
\DeclareAcronym{tcav}{
  short = TCAV,
  long  = Testing Concept Activation Vectors
}
\DeclareAcronym{ubs}{
  short = UBS,
  long  = Uniform unit Ball surface Sampling
}
\DeclareAcronym{mls}{
  short = MLS,
  long  = midline shift
}
\DeclareAcronym{gmm}{
  short = GMM,
  long  = Gaussian mixture model
}
\DeclareAcronym{knn}{
  short = kNN,
  long  = k nearest neighbors 
}

\DeclareAcronym{gru}{
  short = GRU,
  long  =  gated recurrent unit
}
\DeclareAcronym{rnn}{
  short = RNN,
  long  =  recurrent neural network
}
\DeclareAcronym{ehr}{
  short = EHR,
  long  =  electronic healthcare record
}
\DeclareAcronym{ai}{
  short = AI,
  long  =  artificial intelligence
}
\DeclareAcronym{gdpr}{
  short = GDPR,
  long  =  General Data Protection Regulation
}
\DeclareAcronym{hilt}{
  short = HITL,
  long  =  human-in-the-loop
}
\DeclareAcronym{cnv}{
  short =  CNV,
  long  =  choroidal neovascularization
}
\DeclareAcronym{dme}{
  short =  DME,
  long  = diabetic macular edema
}
\DeclareAcronym{oct}{
  short = OCT ,
  long  = optical coherence tomography
}
 
\DeclareAcronym{asd}{
  short = ASD ,
  long  = autism pectrum disorder
}

\begin{document}

\section{Introduction}

\ac{cad} using \ac{ai} provides a promising way to make the diagnosis process more efficient and available to the masses. Deep learning is the leading \ac{ai} method for a wide range of tasks including medical imaging problems. It is the state of the art for several computer vision tasks and has been used for medical imaging tasks like the classification of Alzheimer's \cite{jo2019deep}, lung cancer detection \cite{hua2015computer}, retinal disease detection \cite{sengupta2020ophthalmic, chap_pending}, etc. Despite achieving remarkable results in the medical domain, AI-based methods have not achieved a significant deployment in the clinics. This is due to the underlying black-box nature of the deep learning algorithms along with other reasons like computational costs. It arises from the fact that despite having the underlying statistical principles, there is a lack of ability to explicitly represent the knowledge for a given task performed by a deep neural network. Simpler AI methods like linear regression and decision trees are self-explanatory as the decision boundary used for classification can be visualized in a few dimensions using the model parameters. But these lack the complexity required for tasks such as classification of 3D and most 2D medical images. The lack of tools to inspect the behavior of black-box models affects the use of deep learning in all domains including finance and autonomous driving where explainability and reliability are the key elements for trust by the end-user. 

A medical diagnosis system needs to be transparent, understandable, and explainable to gain the trust of physicians, regulators as well as the patients. Ideally, it should be able to explain the complete logic of making a certain decision to all the parties involved. Newer regulations like the European \ac{gdpr} are making it harder for the use of black-box models in all businesses including healthcare because retraceability of the decisions is now a requirement \cite{holzinger2017we}. An \ac{ai} system to complement medical professionals should have a certain amount of explainability and allow the human expert to retrace the decisions and use their judgment. Some researchers also emphasize that even humans are not always able to or even willing to explain their decisions \cite{holzinger2017we}. Explainability is the key to safe, ethical, fair, and trust-able use of \ac{ai} and a key enabler for its deployment in the real world.  Breaking myths about \ac{ai} by showing what a model looked at while making the decision can inculcate trust among the end-users. It is even more important to show the domain-specific features used in the decision for non-deep learning users like most medical professionals. 

The terms explainability and interpretability are often used interchangeably in the literature. A distinction between these was provided in \cite{stano2020explainable} where interpretation was defined as mapping an abstract concept like the output class into a domain example, while explanation was defined as a set of domain features such as pixels of an image the contribute to the output decision of the model. A related term to this concept is the uncertainty associated with the decision of a model. Deep learning classifiers are usually not able to say "I don't know" in situations with ambiguity and instead return the class with the highest probability, even if by a narrow margin. Lately, uncertainty has been analyzed along with the problem of explainability in many studies to highlight the cases where a model is unsure and in turn make the models more acceptable to non-deep learning users. Deep learning models are considered as non-transparent as the weights of the neurons can't be understood as knowledge directly.  \cite{meyes2020under} showed that neither the magnitude or the selectivity of the activations, nor the impact on network decisions is sufficient for deciding the importance of a neuron for a given task. A detailed analysis of the terminologies, concepts and, use cases of explainable \ac{ai} is provided in \cite{arrieta2020explainable}.

This paper describes the studies related to the explainability of deep learning models in the context of medical imaging. A general taxonomy of explainability approaches is described briefly in the next section and a comparison of various attribution based methods is performed in section \ref{sec:exp_attr}. Section \ref{sec:applications} reviews various explainability methods applied to different medical imaging modalities. The analysis is broken down into subsections \ref{sub:attr} and \ref{sec:non-attr} depending upon the use of attributions or other methods of explainability. The evolution, current trends, and some future possibilities of the explainable deep learning models in medical image analysis are summarized in \ref{sec:discussion}.

\section{Taxonomy of explainability approaches} \label{sec:taxo}
\begin{figure}[hbt!]
\centering
\includegraphics[width=1\linewidth]{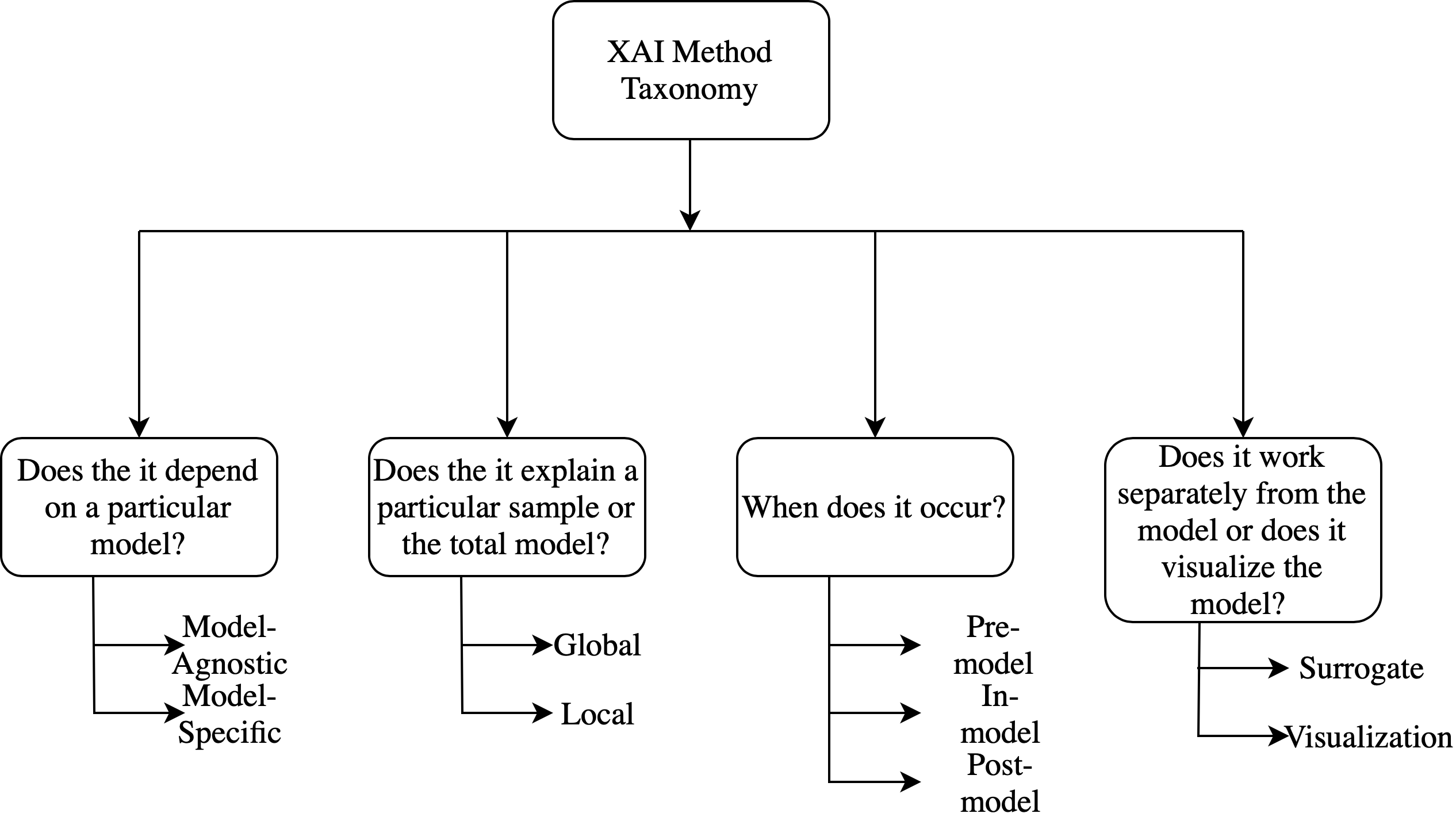}
\caption{Taxonomy of XAI methods} \label{fig:taxonomy}
\end{figure}

Several taxonomies have been proposed in the literature to classify different explainability methods\cite{stiglic2020interpretability,arya2019one}. Generally, the classification techniques are not absolute, it can vary widely depending upon the characteristics of the methods and can be classified into many overlapping or non-overlapping classes simultaneously. Different kinds of taxonomies and classification methods are discussed briefly here and a detailed analysis of the taxonomies can be found in \cite{stiglic2020interpretability, arrieta2020explainable} and a flow chart for them is shown in \ref{fig:taxonomy}.
\subsection{Model Specific vs Model Agnostic}
Model-specific interpretation methods are based on the parameters of the individual models. The graph neural network explainer (GNNExplainer) \cite{ying2019gnnexplainer} is a special type of model-specific interpretability where the complexity of data representation needs specifically the graph neural network (GNN). Model Agnostic methods are mainly applicable in post-hoc analysis and not limited to specified model architecture. These methods do not have direct access to the internal model weights or structural parameters.
\subsection{Global Methods vs Local Methods}
Local interpretable methods are applicable to a single outcome of the model. This can be done by designing methods that can explain the reason for a particular prediction or outcome. For example, it is interested in specific features and their characteristics. On the contrary, global methods concentrate on the inside of a model by exploiting the overall knowledge about the model, the training, and the associated data. It tries to explain the behavior of the model in general. Feature importance is a good example of this method, which tries to figure out the features which are in general responsible for better performance of the model among all different features.
\subsection{Pre-model vs in-model vs post-model}
Pre-model methods are independent and does not depend on a particular model architecture to use it on. Principal component analysis (PCA) \cite{wold1987principal}, t-Distributed Stochastic Neighbor Embedding (t-SNE) \cite{maaten2008visualizing} are some common examples of these methods. Interpretability methods, integrated in the model itself, are called as in-model methods. Some methods are implemented after building a model and hence these methods are termed as post model and these methods can potentially develop meaningful insights about what exactly a model learnt during the training.
\subsection{Surrogate Methods vs Visualization Methods}
Surrogate methods consist of different models as an ensemble which are used to analyze other black-box models. The black box models can be understood better by interpreting the surrogate model’s decisions by comparing the black-box model's decision and surrogate model's decision. The decision tree \cite{safavian1991survey} is an example of surrogate methods. The visualization methods are not a different model, but it helps to explain some parts of the models by visual understanding like activation maps. 

It is to be noted that these classification methods are non-exclusive, these are built upon different logical intuitions and hence have significant overlaps. For example, most of the post-hoc models like attributions can also be seen as model agnostic as these methods are typically not dependent upon the structure of a model. However, some requirements regarding the limitations on model layers or the activation functions do exist for some of the attribution methods. The next section describes the basic concept and subtle difference between various attribution methods to facilitate a comparative discussion of the applications in section \ref{sec:applications}.

\section{Explainability methods - attribution based} \label{sec:exp_attr}

\begin{figure}[hbt!]
\centering
\includegraphics[width=1\linewidth]{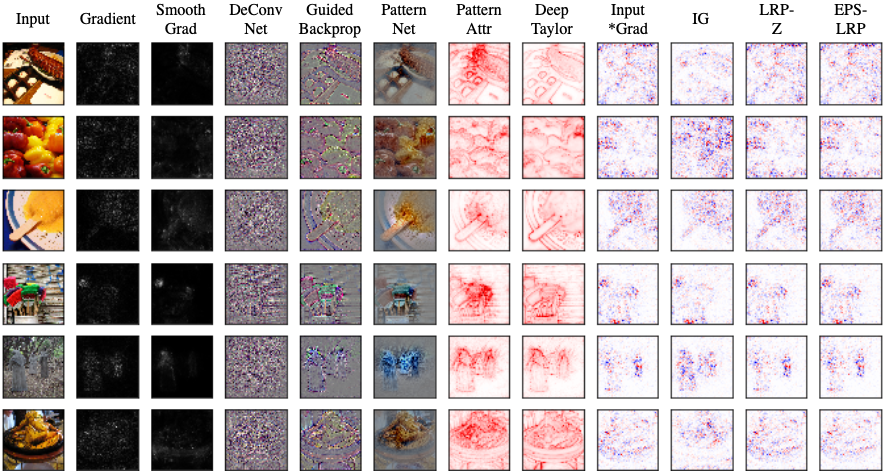}
\caption{Attributions of VGG-16 with images from Imagenet using the methods implemented in \cite{alber2019innvestigate}} \label{fig:innvestigate}
\end{figure}

 There are broadly two types of approaches to explain the results of \ac{dnn} in medical imaging - those using standard attribution based methods and those using novel, often architecture or domain-specific techniques. The methods used for the former are discussed in this section with applications provided in \ref{sub:attr} while the latter are discussed along with their applications in section \ref{sec:non-attr}. The problem of assigning an attribution value or contribution or relevance to each input feature of a network led to the development of several attribution methods. The goal of an attribution method is to determine the contribution of an input feature to the target neuron which is usually the output neuron of the correct class for a classification problem. The arrangement of the attributions of all the input features in the shape of the input sample forms heatmaps known as the \textit{attribution maps}. Some examples of attribution maps for different images are shown in Figure \ref{fig:innvestigate}. The features with a positive contribution to the activation of the target neuron are typically marked in red while those negatively affecting the activation are marked in blue. These are the features or pixels in case of images providing positive and negative evidence of different magnitudes respectively. 
 
The commonly used attribution methods are discussed in this section and the applications in the next section. It must be noted that some of the approaches like DeepTaylor \cite{montavon2017explaining} provide only positive evidence and can be useful for a certain set of tasks.
The attribution methods can be applied on a black box \ac{cnn} without any modification to the underlying architecture making them a convenient yet powerful \ac{xai} tool. An empirical comparison of some of the methods discussed in this section and a unified framework called \textit{DeepExplain} is available in \cite{ancona2017towards}. Most of the methods discussed here apart from the newer Deep \ac{lift} and Deep \ac{shap} are implemented in the iNNvestigate toolbox \cite{alber2019innvestigate}.

\subsection{Perturbation based methods - Occlusion}

Perturbation is the simplest way to analyze the effect of changing the input features on the output of an AI model. This can be implemented by removing, masking, or modifying certain input features, and running the forward pass (output computation), and measuring the difference from the original output. This is similar to the sensitivity analysis performed in parametric control system models. The input features affecting the output the most are ranked as the most important. It is computationally expensive as a forward pass needs to be run after perturbing each group of features of the input. In the case of image data the perturbation is performed by covering parts of an image with a grey patch and hence \textit{occluding} them from the system's view. It can provide both positive and negative evidence by highlighting the responsible features.

This technique was applied by Zeiler and Fergus \cite{zeiler2014visualizing} to the \ac{cnn} for the image classification task. \textbf{Occlusion} is the benchmark for any attribution study as it is a simple to perform model agnostic approach which reveals the feature importance of a model. It can reveal if a model is overfitting and learning irrelevant features as in the case of adversarial examples \cite{goodfellow2014explaining}. The adversarial examples are the inputs designed to cause the model to make a false decision and are like optical illusions for the models. In that case, the model misclassifies the image (say a cat as a dog) despite the presence of discriminating feature

Occluding all features (pixels) one-by-one and running the forward pass each time can be computationally expensive and can take several hours per image \cite{ancona2017towards}. It is common to use patches of sizes such as 5x5, 10x10, or even larger depending on the size of the target features and computational resources available. 

Another perturbation based approach is \textbf{Shapley value sampling} which computes approximate Shapely Values by taking each input feature for a sample number of times. It a method from the coalitional game theory which describes the fair distribution of the gains and losses among the input features. It was originally proposed for the analysis of regression \cite{lipovetsky2001analysis}. It is slower than all other approaches as the network has to be run samples $\times$ number of features times. As a result it is not a practical method in its original form but has led to the development of game theory-based methods like Deep \ac{shap} as discussed in the next subsection.

\subsection{Backpropagation based methods}

\begin{table}[hbtp!]
\begin{center}
\caption{Backpropagation based attribution methods} \label{tab:attr}
\begin{tabular}{|m{0.14\textwidth}|m{0.45\textwidth}|m{0.35\textwidth}|}

\hline \textbf{Method} & \textbf{Description} & \textbf{Notes}  \\ \hline

Gradient & Computes the gradient of the \textbf{output neuron} with respect to the input. & The \textbf{simplest} approach but is usually not the most effective.\\ \hline

DeConvNet \cite{zeiler2014visualizing} & Applies the \textbf{\ac{relu} to the gradient computation instead} of the gradient of a neuron with \ac{relu} activation. & Used to \textbf{visualize the features} learned by the layers. \textbf{Limited} to CNN models with \textbf{\ac{relu} activation}.\\ \hline
 
 Saliency Maps \cite{simonyan2013deep} & Takes the \textbf{absolute value of the partial derivative} of the target output neuron with respect to the input features to find the features which affect the output the most with least perturbation. & \textbf{Can't distinguish between positive and negative} evidence due to absolute values. \\ \hline
 
 \ac{gbp} \cite{springenberg2014striving} & Applies the \textbf{\ac{relu} to the gradient computation in addition} to the gradient of a neuron with \ac{relu} activation.  & Like DeConvNet, it is textbf{limited} to CNN models with \textbf{\ac{relu} activation}.  \\ \hline
 
 \ac{lrp} \cite{bach2015pixel} & \textbf{Redistributes the prediction score} layer by layer with a backward pass on the network using a particular rule like the \textbf{$\epsilon$-rule} while ensuring numerical stability & There are alternative stability rules and \textbf{limited} to CNN models with \textbf{\ac{relu} activation} when all activations are \textbf{\ac{relu}}. \\ \hline
 
 Gradient $\times$ input \cite{shrikumar2016not} & Initially proposed as a method to \textbf{improve sharpness of attribution maps} and is computed by multiplying the signed partial derivative of the output with the input.  & It \textbf{can approximate occlusion} better than other methods in certain cases like \ac{mlp} with Tanh on MNIST data \cite{ancona2017towards} while being instant to compute. \\ \hline
 
 \ac{gradcam} \cite{selvaraju2017grad} & Produces \textbf{gradient-weighted class activation maps} using the gradients of the target concept as it flows to the final convolutional layer & Applicable to \textbf{only CNN} including those with fully connected layers, structured output (like captions) and reinforcement learning. \\ \hline
 
 \ac{ig} \cite{sundararajan2017axiomatic} & Computes the \textbf{average gradient} as the input is varied from the \textbf{baseline} (often zero) to the actual input value unlike the Gradient $\times$ input which uses a single derivative at the input. & It is \textbf{highly correlated with the rescale rule of DeepLIFT} discussed below which can act as a good and faster approximation.\\ \hline
 
 DeepTaylor \cite{montavon2017explaining} & Finds a rootpoint near each neuron with a value close to the input but with output as 0 and uses it to recursively estimate the attribution of each neuron using \textbf{Taylor decomposition} & Provides \textbf{sparser explanations} i.e. focuses on key features but provides \textbf{no negative evidence} due to its assumptions of only positive effect.\\ \hline
 
 PatternNet \cite{kindermans2017learning} &  Estimates the input signal of the output neuron using an \textbf{objective function}. & Proposed to counter the incorrect attributions of other methods on \textbf{linear systems} and generalized to deep networks.\\ \hline
 
 Pattern Attribution \cite{kindermans2017learning}& Applies Deep Taylor decomposition by searching the \textbf{rootpoints in the signal direction} for each neuron & Proposed along with \textbf{PatternNet} and uses decomposition instead of signal visualization\\ \hline
 
 DeepLIFT \cite{shrikumar2017learning} & Uses a reference input and computes the reference values of all hidden units using a forward pass and then proceeds backward \textbf{like \ac{lrp}}. It has two variants - \textbf{Rescale rule} and the one introduced later called \textbf{RevealCancel} which treats positive and negative contributions to a neuron separately. & Rescale is strongly related to and \textbf{equivalent in some cases to $\epsilon$-\ac{lrp}} but is \textbf{not applicable to models involving multiplicative rules}. \textbf{RevealCancel handles such cases} and using RevealCancel for convolutional and Rescale for fully connected layers reduces noise. \\ \hline
 
 SmoothGrad \cite{smilkov2017smoothgrad} & An improvement on the gradient method which averages the gradient over multiple inputs with additional noise & Designed to visually sharpen the attributions produced by gradient method using class score function. \\ \hline
 
 Deep SHAP \cite{chen2019explaining} & It is a fast \textbf{approximation} algorithm to compute the game theory based \textbf{SHAP values}. It is connected to DeepLIFT and uses\textbf{ multiple background samples} instead of one baseline. & Finds attributions for \textbf{non neural net models} like trees, \ac{svm} and \textbf{ensemble} of those with a neural net using various tools in the the SHAP library. \\ \hline
 
\end{tabular}
\end{center}
\end{table}

These methods compute the attribution for all the input features with a single forward and backward pass through the network. In some of the methods these steps need to be repeated multiple times but it is independent of the number of input features and much lower than for perturbation-based methods. The faster run-time comes at the expense of a weaker relationship between the outcome and the variation of the output. Various backpropagation based attribution methods are described in Table \ref{tab:attr}. It must be noted that some of these methods provide only positive evidence while others provide both positive and negative evidence. The methods providing both positive and negative evidence tend to have high-frequency noise which can make the results seem spurious. \cite{ancona2017towards}.

An important property of attribution methods known as \textit{completeness} was introduced in the DeepLIFT \cite{shrikumar2017learning} paper. It states that the attributions for a given input add up to the target output minus the target output at the baseline input. It is satisfied by integrated gradients, DeepTaylor and Deep SHAP but not by DeepLIFT in its rescale rule. A measure generalizing this property is proposed in \cite{ancona2017towards} for a quantitative comparison of various attribution methods. It is called \textit{sensitivity-n} and involves comparing the sum of the attributions and the variation in the target output in terms of \ac{pcc}. Occlusion is found to have a higher \ac{pcc} than other methods as it finds a direct relationship between the variation in the input and that in the output.

The evaluation of attribution methods is complex as it is challenging to discern between the errors of the model and the attribution method explaining it. Measures like sensitivity-n reward the methods designed to reflect the network behavior closely. However, a more practically relevant measure of an attribution method is the similarity of attributions to a human observer's expectation. It needs to be performed with a human expert for a given task and carries an observer bias as the methods closer to the observer expectation can be favored at the cost of those explaining the model behavior. We underscore the argument that the ratings of different attribution methods by experts of a specific domain are potentially useful to develop explainable models which are more likely to be trusted by the end users and hence should be a critical part of the development of an \ac{xai} system.

\section{Applications} \label{sec:applications}

The applications of explainability in medical imaging are reviewed here by categorizing them into two types - those using pre-existing attribution based methods and those using other, often specific methods. The methods are discussed according to the explainability method and the medical imaging application. Table \ref{tab:apps} provides a brief overview of the methods.

\subsection{Attribution based} \label{sub:attr}

A majority of the medical imaging literature that studied interpretability of deep learning methods used attribution based methods due to their ease of use. Researchers can train a suitable neural network architecture without the added complexity of making it inherently explainable and use a readily available attribution model. This allows the use of either a pre-existing deep learning model or one with a custom architecture for the best performance on the given task. The former makes the implementation easier and allows one to leverage techniques like transfer learning \cite{yosinski2014transferable, singh2019glaucoma} while latter can be used to focus on specific data and avoid overfitting by using fewer parameters. Both approaches are beneficial for medical imaging datasets which tend to be relatively smaller than computer vision benchmarks like ImageNet \cite{deng2009imagenet}.

Post-model analysis using attributions can reveal if the model is learning relevant features or if it is overfitting to the input by learning spurious features. This allows researchers to adjust the model architecture and hyperparameters to achieve better results on the test data and in turn a potential real-world setting. In this section, some recent studies using attribution methods across medical imaging modalities such as brain \ac{mri}, retinal imaging, breast imaging, skin imaging, \ac{ct} scans, and chest X-ray are reviewed. \\

\noindent
\textbf{Brain imaging} 

A study comparing the robustness of various attribution based methods for \ac{cnn} in Alzheimer's classification using brain MRI \cite{eitel2019testing} performed a quantitative analysis of different methods. Gradient $\times$ input, \ac{gbp}, \ac{lrp}, and occlusion were the compared methods. The L2 norm between the average attribution maps of multiple runs for the same model to check the repeatability of heatmaps for identically trained models. It was found to be an order of magnitude lower for the first three methods compared to the baseline occlusion since occlusion covers a larger area. LRP performed the best overall indicating the superiority of a completely attribution based method over function and signal-based methods. The similarity between the sum, density, and gain (sum/density) for the top 10 regions of the attributions across the runs was also the highest for LRP. In another study \cite{pereira2018automatic} \ac{gradcam} and \ac{gbp} were used to analyze the clinical coherence of the features learned by a CNN for automated grading of brain tumor from \ac{mri}. For the correctly graded cases, both the methods had the most activation in the tumor region while also activating the surrounding ventricles which can indicate malignancy as well. In some cases, this focus on non-tumor regions and some spurious patterns in \ac{gbp} maps lead to errors indicating unreliability of the features.\\

\noindent
\textbf{Retinal imaging}

\begin{figure}[hbt!]
\centering
\includegraphics[width=1\linewidth]{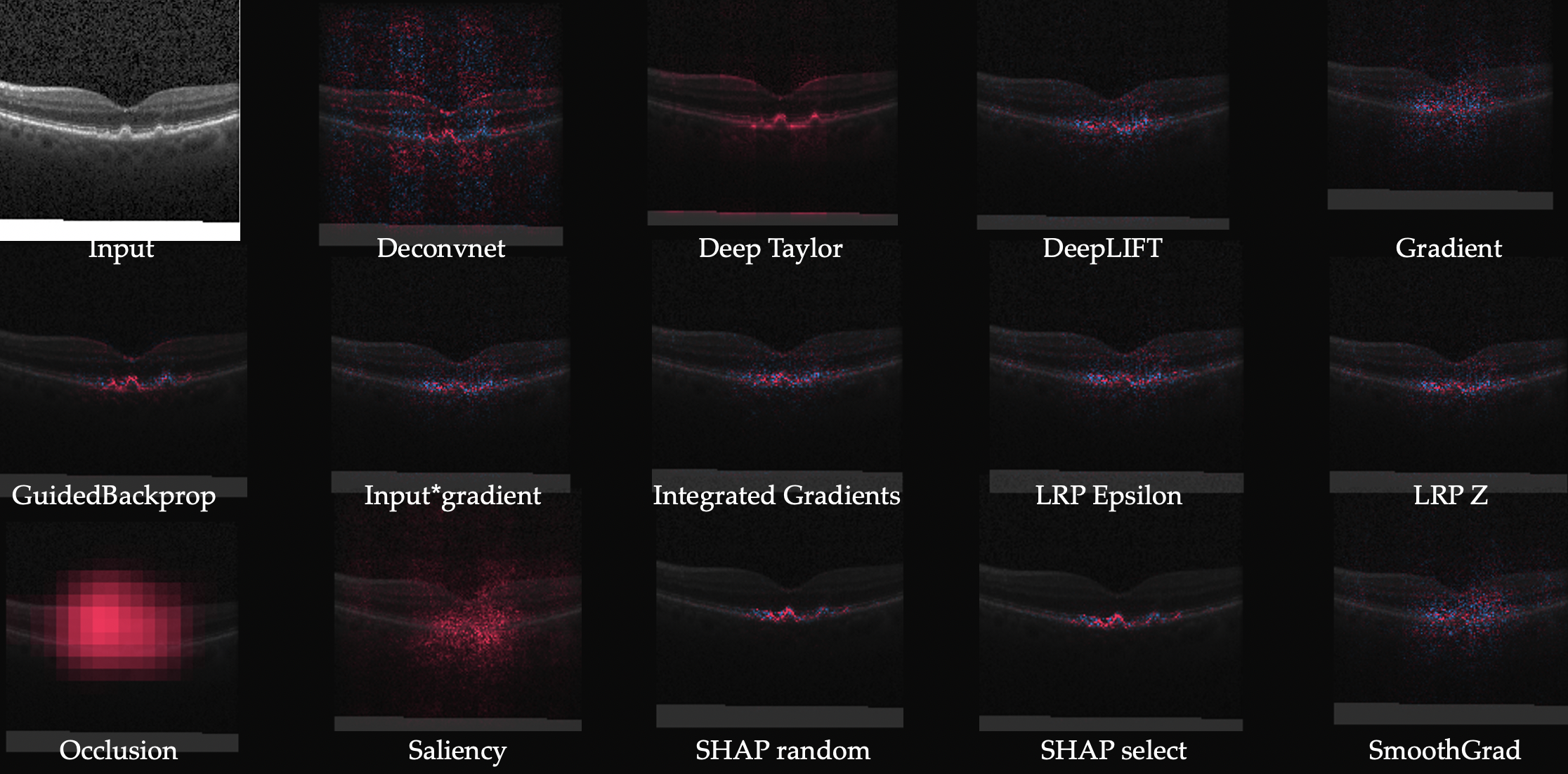}
\caption{Example of heat maps from a retinal OCT image \cite{singh2020interpretation}} \label{fig:oct}
\end{figure}

A system producing \ac{ig} heatmaps along with model predictions was explored as a tool to assist \ac{dr} grading by ophthalmologists \cite{sayres2019using}. This assistance was found to increase the accuracy of the grading compared to that of an unassisted expert or with the model predictions alone. Initially, the system increased the grading time but with the user's experience, the grading time decreased and the grading confidence increased, especially when both predictions and heatmaps were used. Notably, the accuracy did reduce for patients without DR when model assistance was used and an option to toggle the assistance was provided. An extension of IG called \ac{eg} was proposed in \cite{yang2019weakly} for weakly supervised segmentation of lesions for \ac{amd} diagnosis. A \ac{cnn} with a compact architecture outperformed larger existing \ac{cnn}s and \ac{eg} highlighted the regions of interest better than conventional \ac{ig} and \ac{gbp} methods. \ac{eg} extends \ac{ig} by enriching input-level attribution map with high-level attribution maps.A comparative analysis of various explanability models including Deep\ac{lift}, Deep\ac{shap}, \ac{ig}, etc was performed for on a model for detection of \ac{cnv}, \ac{dme}, and drusens from \ac{oct} scans \cite{singh2020interpretation}. Figure \ref{fig:oct} highlights better localization achieved by newer methods (e.g., DeepSHAP) in contrast to noisy results from older methods (e.g., saliency maps).
\\

\noindent
\textbf{Breast imaging}

\ac{ig} and SmoothGrad were used to visualize the features of a \ac{cnn} used for classifying estrogen receptor status from breast MRI \cite{papanastasopoulos2020explainable}. The model was observed to have learned relevant features in both spatial and dynamic domains with different contributions from both. The visualizations revealed the learning of certain irrelevant features resulting from pre-processing artifacts. These observations led to changes in the pre-processing and training approaches. An earlier study for breast mass classification from mammograms \cite{levy2016breast} using two different \ac{cnn}s - AlexNet \cite{szegedy2015going} and GoogleNet \cite{krizhevsky2012imagenet} - employed saliency maps to visualize the image features. Both the \ac{cnn}s were seen to learn the edges of the mass which are the main clinical criteria, while also being sensitive to the context.
\\

\noindent
\textbf{Skin imaging}

The features of a suite of 30 CNN models trained for melanoma detection \cite{young2019deep} were compared using \ac{gradcam} and Kernel \ac{shap}. It was shown that even the models with high accuracy would occasionally focus on the features that were irrelevant for the diagnosis. There were differences in the explanations of the models that produced similar accuracy which was highlighted by the attribution maps of both the methods. This showed that distinct neural network architectures tend to learn different features. Another study \cite{van2018visualizing} visualized the \ac{cnn} features for skin lesion classification. The features for the last two layers were visualized by rescaling the feature maps of the activations to the input size. The layers were observed to be looking at indicators like lesion borders and non-uniformity in color as well as risk factors like lighter skin color or pink texture. However, spurious features like artifacts and hair which have no significance were also learned indicating some extent of overfitting.\\

\noindent
\textbf{CT imaging} 

A DeepDreams \cite{mordvintsev2015inceptionism} inspired attribution method was presented in \cite{couteaux2019towards} for explaining the segmentation of tumor from liver \ac{ct} images. This novel method formulated using the concepts of DeapDreams, an image generation algorithm can be applied to a black-box neural network like other attribution methods discussed in section \ref{sec:exp_attr}. It performed a sensitivity analysis of the features by maximizing the activation of the target neuron by performing gradient ascent i.e. finding the steepest slope of the function. A comparison between networks trained on real tumors and synthetic tumors revealed that the former was more sensitive to clinically relevant features and the latter was focusing on other features too. The network was found to be sensitive to intensity as well as sphericity in coherence with domain knowledge. 
\\

\noindent
\textbf{X-ray imaging}

In a recent study for detection of COVID-19 from chest X-ray images \cite{wang2020covid}, a method called GSInquire was used to produce heatmaps for verifying the features learned by the proposed COVID-net model. GSInquire \cite{lin2019explaining} was developed as an attribution method that outperformed prior methods like SHAP and Expected gradients in terms of the proposed new metrics - impact score and impact coverage. The impact score was defined as the percentage of features which impacted the model decision or confidence strongly. While impact coverage was defined in the context of the coverage of adversarially impacted factors in the input. Another study performed the analysis of uncertainty and interpretability for COVID-19 detection using chest X-rays. The heatmaps of the sample inputs for the trained model were generated using saliency maps, Guided \ac{gradcam}, \ac{gbp}, and \ac{cam}.

There are other studies using attribution based methods for diagnosis in addition to the more common imaging modalities discussed above, both from image and non-image inputs. A study performed uncertainty and interpretability analysis on \ac{cnn}s for semantic segmentation of colorectal polyps, a precursor of rectal cancers \cite{wickstrom2020uncertainty}. Using \ac{gbp} for heatmaps the \ac{cnn}s were found to be utilizing the edge and shape information to make predictions. Also, the uncertainty analysis revealed higher uncertainty in misclassified samples. An explainable model using \ac{shap} attributions for hypoxemia, i.e. low blood oxygen tension prediction during surgery was presented in \cite{lundberg2018explainable}. The study was performed for analyzing preoperative factors as well as in-surgery parameters. The resulting attributions were in line with known factors like BMI, physical status (ASA), tidal volume, inspired oxygen, etc.

The attribution based methods were one of the initial ways of visualizing neural networks and have since then evolved from simple class activation map and gradient-based methods to advanced techniques like Deep \ac{shap}. The better visualizations of these methods show that the models were learning relevant features in most of the cases. Any presence of spurious features was scrutinized, flagged to the readers, and brought adjustments to the model training methods. Smaller and task-specific models like \cite{yang2019weakly} along with custom variants of the attribution methods can improve the identification of relevant features.

\subsection{Non-attribution based} \label{sec:non-attr}

The studies discussed in this subsection approached the problem of explainability by developing a methodology and validating it on a given problem rather than performing a separate analysis using pre-existing attributions based methods like those previously discussed. These used approaches like attention maps, concept vectors, returning a similar image, text justifications, expert knowledge, generative modeling, combination with other machine learning methods, etc. It must be noted that the majority of these are still post-model but their implementation usually needs specific changes to the model structure such as in the attention maps or the addition of expert knowledge in case of rule-based methods. In this section, the studies are grouped by the explainability approach they took.

\subsubsection{Attention based}
Attention is a popular and useful concept in deep learning. The basic idea of attention is inspired by the way humans pay attention to different parts of an image or other data sources to analyze them. More details about attention mechanisms in neural networks is discussed in \cite{vaswani2017attention}. An example of attention in medical diagnosis is given in \cite{bamba2020classification}. Here, we discuss how attention-based methods can be used as an explainable deep learning tool for medical image analysis.\\
A network called MDNet was proposed \cite{zhang2017mdnet} to perform a direct mapping between medical images and corresponding diagnostic reports. With an image model and a language model in it, the method used attention mechanisms to visualize the detection process. Using that attention mechanism, the language model found predominant and discriminatory features to learn the mapping between images and the diagnostic reports. This was the first work which exploited the attention mechanism to get insightful information from medical image dataset.\\
In \cite{sun2020saunet} an interpretable version of U-Net \cite{ronneberger2015u} called SAUNet was proposed. It added a parallel secondary shape stream to capture important shape-based information along with the regular texture features of the images. The architecture used an attention module in the decoder part of the U-Net. The spatial and shape attention maps were generated using SmoothGrad to visualize the high activation region of the images.

\subsubsection{Concept vectors} \label{concept}

A novel method called \ac{tcav} was proposed in \cite{kim2018tcav} to explain the features learned by different layers to the domain experts without any deep learning expertise in terms of human-understandable concepts. It took the directional derivative of the network in the concept space much like that in the input feature space for saliency maps. It was tested to explain the predictions of \ac{dr} levels where it successfully detected the presence of microaneurysms and aneurysms in the retina. This provided justifications that were readily interpretable for the medical practitioners in terms of presence or absence of a given concept or physical structure in the image. However, many clinical concepts like the texture or the shape of a structure cannot be sufficiently described in terms of the presence or absence and need a continuous scale of measurement.

 An extension of \ac{tcav}, which used the presence or absence of concepts, using \ac{rcv} in the activation space of a layer was used to detect continuous concepts \cite{graziani2018regression}. IThe task of the network was to detect tumors from breast lymph node samples. It was found that most of the relevant features like area and contrast were present in the early layers of the model. A further improvement over the \ac{tcav} used a new metric called \ac{ubs} \cite{yeche2019ubs} to provide layer-agnostic explanations for continuous and high dimensional features. It could explain high dimensional radiomics concepts across multiple layers which were validated using mammographic images. The model produced variations amongst the important concepts which were found to be lower across the layers of the SqueezeNet \cite{iandola2016squeezenet} compared to a baseline CNN with 3 dense layers explaining the better performance of the SqueezeNet.

\subsubsection{Expert knowledge}

A vast majority of the research discussed in this review tried to correlate model features with expert knowledge using different approaches. Another approach was to use domain-specific knowledge to craft rules for prediction and explanation. An example of using task-specific knowledge to improve the results as well as the explanations were provided in \cite{pisov2019incorporating} for brain \ac{mls} estimation using U-Net \cite{ronneberger2015u} based architecture and keypoints. It was reduced to the problem of detecting a midline using the model under domain constraints. The original midline was obtained using the endpoints and hence the shift from the predicted one was computed. The model also provided confidence intervals of the predictions making them more trustworthy for the end-user. Another study \cite{zhu2019guideline} used guidelines for rule-based segmentation of lung nodules followed by a perturbation analysis to compute the importance of features in each region. The explanations provided in terms of the regions already marked using rules were found to be more understandable for the users and showed the bais in data for improving the model. This method was then used to provide explanations at a global level for the entire dataset providing an overview of the relevant features.

\subsubsection{Similar images}

Some studies provided similarly labeled images to the user as a reason for making a prediction for a given test image. A study \cite{stano2020explainable} proposed analysis of layers of a 3D-\ac{cnn} using \ac{gmm} and binary encoding of training and test images based on their \ac{gmm} components for returning similar 3D images as explanations. The system returned activation wise similar training images using atlas as a clarification for its decision. It was demonstrated on 3D MNIST and an \ac{mri} dataset where it returned images with similar atrophy conditions. However, it was found that the activation similarity depended on the spatial orientation of images in certain cases which could affect the choice of the returned images. 

In a study on dermoscopic images, a triplet-loss and \ac{knn} search-based learning strategy was used to learn \ac{cnn} feature embeddings for interpretable classification \cite{codella2018collaborative}. The evidence was provided as nearest neighbors and local image regions responsible for the lowest distance between the test image and those neighbors. Another approach used monotonic constraints to explain the predictions in terms of style and depth two datasets - dermoscopy images and post-surgical breast aesthetics \cite{silva2018towards}. It concatenated input streams with constrained monotonic \ac{cnn} and unconstrained \ac{cnn} to produce the predictions along with their explanations in terms of similar images as well as complementary images. The system was designed for only binary classification. 

\subsubsection{Textual justification}

A model that can explain its decision in terms of sentences or phrases giving the reasoning can directly communicate with both expert and general users. A justification model that took inputs from the visual features of a classifier, as well as embeddings of the predictions, was used to generate a diagnostic sentence and visual heatmaps for breast mass classification \cite{lee2019generation}. A visual word constraint loss was applied in the training of the justification generator to produce justifications in the presence of only a limited number of medical reports. Such multimodal explanations can be used to obtain greater user confidence due to a similarity with the usual workflow and learning process.

\subsubsection{Intrinsic explainability}

Intrinsic explainability refers to the ability of a model to explain its decisions in terms of human observable decision boundaries or features. These usually include relatively simpler models like regression, decision trees and \ac{svm} for a few dimensions where the decision boundaries can be observed. Recent studies to make deep learning model intrinsically explainable using different methods such as a hybrid with machine learning classifiers and visualizing the features in a segmentation space. 

An example of the latter was presented in \cite{biffi2020explainable} using the latent space of the features of a variational autoencoder for classification and segmentation of the brain \ac{mri} of Alzheimer's patients. The classification was performed in a two-dimensional latent space using an \ac{mlp}. The segmentation was performed in a three-dimensional latent space in terms of the anatomical variability encoded in the discriminating features. This led to the visualization of the features of the classifier as global and local anatomical characteristics which were usually used for clinical decisions. A study for detection of \ac{asd} from \ac{fmri} used a hybrid of deep learning and \ac{svm} to perform explainable classification \cite{eslami2020explainable}. The \ac{svm} was used as a classifier on the features of a deep learning model and the visualization of the decision boundary explained the model.

This subsection discussed a variety of non-attribution explainability methods but the list is not exhaustive as newer methods are published frequently due to high interest in the area. The design of these methods is more involved than the application of attribution based methods on the inputs of a trained model. Specific elements like concept vectors, expert-based rules, image retrieval methods need to be integrated often at a model training level. This added complexity can potentially provide more domain-specific explanations at the expense of higher design effort. Notably, a majority of these techniques are still a post-hoc step but for a specific architecture or domain.  Also, we have limited our scope to medical imaging as that is the dominant approach for automated diagnosis because of the detailed information presented by the images. However, patient records also provide rich information for diagnosis and there were studies discussing their explainability. For example, in \cite{sha2017interpretable} a \ac{gru}-based \ac{rnn} for mortality prediction from diagnostic codes from \ac{ehr} was presented. It used hierarchical attention in the network for interpretability and visualization of the results.

\begin{table}[hbtp!]

\begin{center}
\caption{Applications of explainability in medical imaging}\label{tab:apps}
\begin{tabular}{|m{0.13\textwidth}|m{0.22\textwidth}|m{0.15\textwidth}| m{0.22\textwidth} | m{0.14\textwidth} |} 

\hline

\textbf{Method} &  \textbf{Algorithm} & \textbf{Model} & \textbf{Application} & \textbf{Modality}  \\ \hline
\multirow{9}{4em}{Attribution} & Gradient*I/P, GBP, LRP, occlusion \cite{eitel2019testing} & 3D CNN & Alzheimer's detection & Brain MRI\\ \cline{2-5} 
 & GradCAM, GBP \cite{pereira2018automatic} & Custom CNN & Grading brain tumor & Brain MRI\\ \cline{2-5}
 & IG \cite{sayres2019using} & Inception-v4 & DR grading & Fundus images\\\cline{2-5}
 & EG \cite{yang2019weakly} & Custom CNN & Lesion segmentation for AMD & Retinal OCT\\ \cline{2-5}
 & IG, SmoothGrad \cite{papanastasopoulos2020explainable} & AlexNet & Estrogen receptor status & Breast MRI \\ \cline{2-5}
 & Saliency maps \cite{levy2016breast} & AlexNet & Breast mass classification & Breast MRI \\ \cline{2-5}
 & GradCAM, SHAP \cite{young2019deep} & Inception & Melanoma detection & Skin images \\ \cline{2-5}
 & Activation maps \cite{van2018visualizing} & Custom CNN & 
Lesion classification & Skin images \\\cline{2-5}
 & DeepDreams \cite{couteaux2019towards} & Custom CNN & Segmentation of tumor from liver & CT imaging \\ \cline{2-5}
 & GSInquire, GBP, activation maps \cite{wang2020covid} & COVIDNet CNN & COVID-19 detection & X-ray images \\\cline{2-5}\hline
 \multirow{2}{4em}{Attention} &  Mapping between image to reports \cite{zhang2017mdnet} & CNN \& LSTM & Bladder cancer & Tissue images \\\cline{2-5}
  &  U-Net with shape attention stream \cite{sun2020saunet} & U-net based & Cardiac volume estimation  & Cardiac MRI \\\hline
   \multirow{3}{4em}{Concept vectors} & TCAV \cite{kim2018tcav} & Inception & DR detection & Fundus images\\\cline{2-5}
  & TCAV with RCV \cite{graziani2018regression} & ResNet101 & Breast tumor detection & Breast lymph node images\\\cline{2-5}
  & UBS \cite{yeche2019ubs} & SqueezeNet & Breast mass classification & Mammography images \\ \hline
 \multirow{2}{4em}{Expert knowledge} & Domain constraints \cite{pisov2019incorporating} & U-net & Brain MLS estimation & Brain MRI  \\ \cline{2-5}
  & Rule-based segmentation, perturbation \cite{zhu2019guideline} & VGG16 &  Lung nodule segmentation & Lung CT \\ \hline
 \multirow{2}{4em}{Similar images} & GMM and atlas \cite{stano2020explainable} & 3D CNN & MRI classification & 3D MNIST, Brain MRI \\\cline{2-5}
  & Triplet loss, kNN \cite{codella2018collaborative} & AlexNet based with shared weights & Melanoma & Dermoscopy images \\\cline{2-5}
  & Monotonic constraints \cite{silva2018towards} & DNN with two streams & Melanoma detection & Dermoscopy images\\ \hline
 Textual justification & LSTM, visual word constraint \cite{lee2019generation}  & Breast mass classification & CNN & Mammography images \\ \hline
 \multirow{2}{4em}{Intrinsic explainability} & Deep Hierarchical Generative Models \cite{biffi2020explainable} & Auto-encoders & Classification and segmentation for Alzheimer's & Brain MRI \\\cline{2-5}
  & SVM margin \cite{eslami2020explainable} & Hybrid of CNN \& SVM & ASD detection & Brain fMRI \\\cline{2-5}\hline
 \end{tabular}
\end{center}
\end{table}

\section{Discussion} \label{sec:discussion}

There has been significant progress in explaining the decisions of deep learning models, especially those used for medical diagnosis. Understanding the features responsible for a certain decision is useful for the model designers to iron out reliability concerns for the end-users to gain trust and make better judgments. Almost all of these methods target local explainability, i.e. explaining the decisions for a single example. This then is extrapolated to a global level by averaging the highlighted features, especially in cases where the images have the same spatial orientation. However, emerging methods like concept vectors (\ref{concept}) provide a more global view of the decisions for each class in terms of domain concepts. 

It is important to analyze the features of a black-box which can make the right decision due to the wrong reason. It is a major issue that can affect performance when the system is deployed in the real world. Most of the methods, especially the attribution based are available as open source implementations. However, some methods like GSInquire \cite{lin2019explaining} which show higher performance on some metrics are proprietary. There is an increasing commercial interest in explainability, and specifically the attribution methods which can be leveraged for a variety of business use cases.

Despite all these advances, there is still a need to make the explainability methods more holistic and interwoven with uncertainty methods. More studies like \cite{sayres2019using} need to be conducted to observe the effect of the explainability models on the decision time and accuracy of the clinical experts. Expert feedback must be incorporated into the design of such explainability methods to tailor the feedback for their needs. Initially, any clinical application of such explainable deep learning methods is likely to be a \ac{hilt} hybrid keeping the clinical expert in the control of the process. It can be considered analogous to driving aids like adaptive cruise control or lane keep assistance in cars where the driver is still in control and responsible for the final decisions but with a reduced workload and an added safety net. 

Another direction of work can be to use multiple modalities like medical images and patients' records together in the decision-making process and attribute the model decisions to each of them. This can simulate the diagnostic workflow of a clinician where both images and physical parameters of a patient are used to make a decision. It can potentially improve the accuracy as well as explain in a more comprehensive way. To sum it up, explainable diagnosis is making convincing strides but there is still some way to go to meet the expectations of end-users, regulators, and the general public.


\funding{This research was supported by a Discovery Grant from NSERC, Canada to Vasudevan Lakshminarayanan.}


\conflictsofinterest{The authors declare no conflict of interest.} 


\printacronyms

\reftitle{References}
\bibliography{main.bib}


\end{document}